\documentclass[letterpaper, 10 pt, conference]{ieeeconf}  

\IEEEoverridecommandlockouts                              

\usepackage{subcaption}
\usepackage{xcolor,soul,framed} 
\colorlet{shadecolor}{yellow}
\usepackage[pdftex]{graphicx}
\graphicspath{{../pdf/}{../jpeg/}}
\DeclareGraphicsExtensions{.pdf,.jpeg,.png}
\usepackage[cmex10]{amsmath}
\usepackage{array}
\usepackage{mdwmath}
\usepackage{mdwtab}
\usepackage{eqparbox}
\usepackage{url}
\usepackage{amsmath}
\usepackage{amssymb}
\usepackage{mathrsfs}
\usepackage{graphicx}
\usepackage{algorithm}
\usepackage{algpseudocode}
\usepackage{booktabs}
\usepackage{multirow}
\usepackage{bm}
\usepackage{xcolor}
\usepackage{float}
\usepackage{array}    
\usepackage{rotating} 
\usepackage{verbatim}
\usepackage{svg}
\usepackage{cite}
\usepackage{caption}

\title{\LARGE \bf
Correspondence-Free Multiview Point Cloud Registration via Depth-Guided Joint Optimisation 
}

\author{Yiran Zhou$^{1}$, Yingyu Wang$^{1}$, Shoudong Huang$^{1}$ and Liang Zhao$^{2}$
  \thanks{Yiran Zhou, Yingyu Wang and Shoudong Huang are with Robotics Institute, Faculty of Engineering and Information Technology, University of Technology Sydney, Sydney, Australia (e-mail: Yiran.Zhou-1@student.uts.edu.au; Yingyu.Wang-1@student.uts.edu.au; Shoudong.Huang@uts.edu.au)}
  \thanks{Liang Zhao is with the School of Informatics, The University of Edinburgh, Edinburgh, UK (e-mail: Liang.Zhao@ed.ac.uk).}
}

\begin{document}

\maketitle
\thispagestyle{empty}
\pagestyle{empty}

\begin{abstract}

Multiview point cloud registration is a fundamental task for constructing globally consistent 3D models. Existing approaches typically rely on feature extraction and data association across multiple point clouds; however, these processes are challenging to obtain global optimal solution in complex environments. In this paper, we introduce a novel correspondence-free multiview point cloud registration method. Specifically, we represent the global map as a depth map and leverage raw depth information to formulate a non-linear least squares optimisation that jointly estimates poses of point clouds and the global map. Unlike traditional feature-based bundle adjustment methods, which rely on explicit feature extraction and data association, our method bypasses these challenges by associating multi-frame point clouds with a global depth map through their corresponding poses. This data association is implicitly incorporated and dynamically refined during the optimisation process. Extensive evaluations on real-world datasets demonstrate that our method outperforms state-of-the-art approaches in accuracy, particularly in challenging environments where feature extraction and data association are difficult. 

\end{abstract}

\section{INTRODUCTION}

 Point cloud registration is a fundamental task with extensive applications across various domains, including 3D reconstruction, odometry estimation, multi-sensor point cloud fusion, augmented reality, and virtual reality. 3D point clouds are generated from depth data captured by sensors or reconstructed through stereo image matching. Since data is acquired from different sensors or at different times, point clouds are represented in distinct local coordinates. The goal of point cloud registration is to align these disparate point clouds into a unified coordinate system, facilitating the construction of a consistent and comprehensive 3D model. 

The most common approach to point cloud registration is pairwise registration, which estimates the transformation needed to align a source point cloud with a target point cloud. However, in applications such as 3D reconstruction, a single pair of point clouds is often insufficient to capture the complete structure of an object. To address this, multiple point clouds must be aligned within a unified coordinate system, a process known as multiview point cloud registration. 

Early solutions to the multiview registration problem relied on sequential pairwise registration\cite{faugeras1986representation}. However, such approaches suffer from accumulating relative pose errors as the number of frames increases, ultimately failing to produce a globally consistent point cloud. To mitigate this issue, a common approach is to formulate the problem as pose graph optimisation (PGO) \cite{mendes2016icp,wang2023robust} or feature-based bundle adjustment (BA) \cite{triggs2000bundle,dellaert2006square,zhao2015parallaxba,liu2021balm,liu2023efficient,huang2021bundle}. PGO-based methods optimise only the poses, while feature-based BA approaches jointly refine both poses and features in the global map. Typically, the joint optimization scheme can lead to more accurate results. 

The feature-based BA method for multiview registration typically consists of two key steps. The first step, data association, involves identifying correspondences between features across multiple point clouds. Then, using these known correspondences, the problem is formulated as a nonlinear least squares (NLLS) optimisation, relating both poses and features. When data association is accurate, the NLLS problem can be effectively solved using reliable nonlinear optimisation solvers \cite{agarwal2012ceres,dellaert2012factor}. However, it is a major challenge to establish reliable data association in complex environments, such as those with highly repetitive textures.

In this paper, we propose a novel correspondence-free multiview point cloud registration method guided by depth information. We leverage the depth information of each point cloud to establish constraints on the global point cloud and poses of individual point clouds. Specifically, we formulate multiview registration as a joint optimisation problem, treating both poses of all point cloud frames and the global map as variables. The key novelty of our approach lies in representing the global map as a depth map, where each point in point clouds is associated with the global depth map via its corresponding poses. In this formulation, data association between multi-frame point clouds is implicitly incorporated through the projection relationship between the point clouds and the depth map and dynamically refined during the optimisation process. Our experimental results demonstrate that the proposed method outperforms state-of-the-art approaches in real-world scenarios, particularly in challenging environments where establishing accurate data association is difficult. The main contributions of our work are summarised as follows: 

\begin{enumerate}
\item We formulate multiview point cloud registration as a joint optimisation problem, which simultaneously optimises the poses of multiple point clouds and the global map.
\item We represent the global map as a depth map, leveraging raw depth information to guide the optimisation process. This eliminates the need for explicit data association, enabling robustness in complex environments. 
\item We provide an analytical derivation of the Jacobian for the proposed optimisation problem, ensuring efficient and accurate problem-solving. 
\item Extensive experiments on real-world datasets demonstrate that our method outperforms state-of-the-art approaches in robustness and accuracy. 
\end{enumerate}

\section{RELATED WORK}

\subsection{Pairwise Registration}
Pairwise registration is a foundational approach in point cloud registration, widely adopted due to its simplicity and effectiveness in aligning two scans. The earliest method, Iterative Closest Point (ICP) \cite{besl1992method}, iteratively minimises point-to-point distance but highly sensitive to initial estimates, noise, and outliers, often leading to local minima. To improve robustness and convergence, variants like point-to-plane ICP \cite{chen1992object}, Generalised ICP \cite{segal2009generalized}, and Anisotropic ICP \cite{pavlov2018aa} incorporate local point cloud structures and anisotropic covariance modelling. However, these methods remain sensitive to initial pose estimates and struggle in noisy or sparse environments. 

To enhance the robustness of these methods, RANSAC \cite{bustos2017guaranteed} and M-estimator \cite{li2024partial} are commonly employed. However, its performance deteriorates as the outlier ratios increase. Recently, TEASER \cite{yang2020teaser}, a state-of-the-art sequential method for point cloud registration, has been proposed. It follows a sequential approach by estimating pairwise transformations between consecutive frames and incrementally constructing the global point cloud. By leveraging a truncated least-squares optimisation framework, TEASER enhances robustness and efficiency, making it particularly effective under extreme outlier conditions.

Although pairwise registration methods can efficiently and robustly estimate the relative transformation between two point clouds, a single two-frame alignment is often insufficient to capture the complete structure of an object or environment. This limitation restricts their applicability in tasks such as 3D reconstruction.

\subsection{Multiview Registration}
Multiview registration aims to optimise the poses of multiple overlapping point cloud frames simultaneously to achieve global consistency. Unlike pairwise registration methods which only consider two frames at a time, multiview registration addresses the global alignment of all frames, reducing accumulated errors and improving overall accuracy. 

PGO methods \cite{lu1997globally,pulli1999multiview} have emerged as a popular approach for multiview registration due to their computational efficiency and scalability. By representing the environment as a graph of keyframes connected by relative pose constraints, these methods efficiently optimise camera trajectories while maintaining real-time performance. However, PGO only optimises the poses of point cloud frames while neglecting the map. This approximation may contribute to suboptimal detail preservation in the registered 3D global point cloud.

In contrast, BA-based methods can jointly optimise poses and maps, fully utilizing observation information. As a result, they are expected to achieve higher precision in reconstruction and more accurate camera pose estimation than PGO-based methods. Traditional feature-based BA methods \cite{triggs2000bundle} rely on extracting feature points and establishing constraints by identifying the same feature points across multiple point clouds. However, despite extensive research on feature points extraction \cite{rusu2008aligning,tombari2013performance}, reliably detecting and accurately associating feature points in complex scenes with repetitive textures remains challenging, limiting the effectiveness of traditional feature-based BA. 

To overcome these limitations, recent research has explored parameterising point clouds into geometric features (e.g., planes or edges) for BA formulation. For example, BALM \cite{liu2021balm} leverages geometric primitives (e.g., planes, lines) to enhance optimisation stability and reduce computational complexity, while BALM2 \cite{liu2023efficient} and BAREG \cite{huang2021bundle} further improve efficiency by using point clusters and avoiding individual point enumeration. These methods enable the joint optimisation of poses and geometric features, improving accuracy in structured environments. However, their performance deteriorates in unstructured scenes where distinct geometric features are scarce. 

Therefore, existing multiview point cloud registration typically depends on explicit feature extraction and data association, which can be unreliable in feature-sparse, noisy environments and unstructured scenes.

\subsection{Joint Optimisation of Poses and Non-feature Map}

Several other BA methods also avoid explicit feature extraction and data association by jointly optimising both pose and non-feature-based maps. BAD-SLAM \cite{schops2019bad} employs a direct BA approach that minimises both reprojection photometric errors and geometric errors. However, directly optimising all pixel points is computationally expensive. To mitigate this, BAD-SLAM adopts an approximation scheme that first optimises only the pose and then updates the surfel representation, which inevitably reduces optimisation accuracy. Additionally, Occupancy-SLAM \cite{Zhao-RSS-22,wang2025occupancy} jointly optimises robot poses and an occupancy grid map to enhance localisation and mapping accuracy. Similarly, \cite{wang2024grid} introduces an efficient framework to optimise a global occupancy map alongside the coordinate frames of local submaps. Kimera-PGMO \cite{rosinol21ijrr-kimera} jointly optimises poses and mesh representations. While these methods also consider joint optimisation of poses and non-feature-based maps, they are not specifically designed for multiview point cloud registration and employ different non-feature map representations as compared with this paper. 

\section{METHODOLOGY}
Our approach considers the joint optimisation of the camera poses and the depth map, leveraging raw depth information without the need for explicit data association. In this section, we will illustrate how the depth observations can be linked to the camera poses to formulate the NLLS problem. Additionally, we obtained an analytical Jacobian derived from the gradient of the depth map, ensuring efficient and accurate optimisation.

\subsection{Task of Multiview Registration and Depth Constraint}\label{sec_31}

The input to multiview point cloud registration is a sequence of point clouds, denoted as $\bm{P} = \{\bm{P}_i \mid i = 1, \ldots, N\}$. The task is to estimate their corresponding poses $\bm{X}^r = \{ \bm{X}_i^r \in SE(3) \mid i= 1, \ldots, N\} $, where $\bm{X}_i^r = \left[\bm{t}_i^\top, \bm{ \theta}^\top_i\right]^\top$, so that a consistent global 3D point cloud can be obtained by projecting the individual point clouds into the global coordinate system using these estimated poses. Here, $\bm{t}_i = \left[t_i^x, t_i^y, t_i^z \right]^\top $ represents x-y-z position, and $\theta_i = \left[\theta_i^x, \theta_i^y, \theta_i^z\right]^\top$ represents the Euler angles (roll, pitch, yaw) corresponding to rotation matrix $\bm{R}_i$. 

To model the global environment, we first represent it as a 2D depth map $\bm{D} = [ \cdots, \bm{D}(\bm{d}_{m,n}), \cdots]$ in the global coordinate, where $\bm{D}(\bm{d}_{m,n})$ represents the depth of grid $\bm{d}_{m,n} (1\leq m \leq l_m, 1\leq n \leq l_n)$. The the depth map resolution, $s$, represents the real-world distance between adjacent grids. The depth value of each grid $\bm{D}(\bm{d}_{m,n})$ in the depth map is computed by averaging the depth values of all the points from the 3D point cloud whose x-y coordinates lie within this grid.

Next, the position of the $j$-th point in the $i$-th point cloud $\bm{P}_i$, denoted as $\bm{p}_{ij}$, can be projected into the global coordinate using its corresponding pose \(\bm{X}_{i}^r\), i.e.,
\begin{equation}
\bm{p'}_{ij} ={\bm{R}_i\bm{p}_{ij} + \bm{t}_i} = [x_{ij}',y_{ij}',z_{ij}']^\top.\label{eq_proj}
\end{equation}

Assuming that the poses are accurate and the environment is static, the depth value of the projected point located on depth map $\bm{D}([\bm{p'}_{ij}]_{xy}/s)$ should be very closed to its depth measurement in the global coordinate $[\bm{p'}_{ij}]_z$, which forms the depth constraint. Here, 
\begin{equation}
    [\bm{p'}_{ij}]_{xy} = [x'_{ij},y'_{ij}]^\top, \quad
    [\bm{p'}_{ij}]_z = z'_{ij}. \label{eq_xyz}
\end{equation}
The depth measurement $[\bm{p'}_{ij}]_z$ can be obtained directly from depth sensors (e.g., LiDAR, structured light, and depth cameras) or estimated via a stereo matching algorithm when using a stereo camera, and then projected into the global coordinate.

\subsection{NLLS Formulation} \label{sec_NLLS}
Based on the depth constraint described in Section \ref{sec_31}, we now formulate the NLLS problem to simultaneously optimise the robot poses and the depth map. The state vector of the proposed problem is
\begin{equation}
	\bm{X} = [(\bm{X}^r)^\top, \bm{D}^\top]^\top,
\end{equation}
where 
\begin{equation}
\begin{aligned}
\bm{X}^r & =\left[\left(\bm{X}_1^r\right)^\top, \cdots,\left(\bm{X}_N^r\right)^\top\right]^\top, \\
\bm{D} & =\left[\bm{D}\left({\bm{d}}_{1,1}\right), \cdots, \bm{D}\left({\bm{d}}_{l_m,l_n}\right)\right]^\top.
\end{aligned}
\end{equation}

The objective function is defined as
\begin{equation}
f(\bm{X}) = w_D f^D(\bm{X}) + w_S f^S(\bm{X}),
\label{NLLS}
\end{equation}
where $f^D(\bm{X})$ and $f^S(\bm{X})$ represent the depth constraint term and the smoothing term, respectively. $ w_D$ and $ w_S$ denotes their corresponding weights.
\subsubsection{Depth Constraint Term}

By the local-to-global projection in (\ref{eq_proj}), all points in the point cloud sequence $\bm{P}$ can be projected to the depth map $\bm{D}$ to compute the difference in depth values to formulate the NLLS problem, i.e., minimise
\begin{equation}
f^D(\bm{X}) = \sum_{i=1}^N \sum_{j} \left\| [\bm{p'}_{ij}]_z - \bm{D}\left(\frac{[\bm{p'}_{ij}]_{xy}}{s}\right) \right\|^2.
\end{equation}

Since $[\bm{p'}_{ij}]_{xy}/s$ may lie at any position on depth map $\bm{D}$ rather than on a discretised grid, its depth value $ \bm{D}([\bm{p'}_{ij}]_{xy}/s)$ can be approximated by bilinear interpolation using depth values of the four neighbouring grids around it. Suppose $[\bm{p'}_{ij}]_{xy}/s$ locates within four grids, $\bm{d}_{m,n},\bm{d}_{m+1,n},\bm{d}_{m,n+1},\bm{d}_{m+1,n+1}$, the depth value of $[\bm{p'}_{ij}]_{xy}/s$ can be calculated by
\begin{equation}
\bm{D}\left(\frac{[\bm{p'}_{ij}]_{xy}}{s}\right) = \bm{H} \left[ \begin{array}{l}
\bm{D}(\bm{d}_{m,n}) \\
\bm{D}(\bm{d}_{m+1,n}) \\
\bm{D}(\bm{d}_{m,n+1}) \\
\bm{D}(\bm{d}_{m+1,n+1})
\end{array} \right]
\end{equation}
where $\bm{H}$ denotes the bilinear interpolation coefficients. 

By using the depth constraint term, the sequence of point clouds $\bm{P}$, their corresponding poses $\bm{X}^r$, and the depth map $\bm{D}$ are linked together.

\subsubsection{Smoothing Term}

To improve the robustness and convergence of our method, we introduce a smoothing term by penalising large variations between neighbouring grids, ensuring that depth transitions smoothly across the map, i.e.,
\begin{equation}
\begin{aligned}
f^S(\bm{X})&= \sum_{m=1}^{l_{m}-1} \sum_{n=1}^{l_{n}-1} \left\| \begin{bmatrix}
           \bm{D}(\bm{d}_{m,n}) - \bm{D}(\bm{d}_{m+1,n}) \\
           \bm{D}(\bm{d}_{m,n}) - \bm{D}(\bm{d}_{m,n+1})
       \end{bmatrix} \right\|^2 \\
       &\quad + \sum_{n=1}^{l_n-1} \left\| \bm{D}(\bm{d}_{l_{m},n}) - \bm{D}(\bm{d}_{l_{m},n+1}) \right\|^2 \\
       &\quad + \sum_{m=1}^{l_m-1} \left\| \bm{D}(\bm{d}_{m,l_{n}}) - \bm{D}(\bm{d}_{m+1,l_{n}}) \right\|^2 .
\end{aligned} 
\end{equation}

\subsection{Iterative Solution and Analytical Jacobian}
\subsubsection{Iterative Solution}
Our NLLS formulation in (\ref{NLLS}) seeks $\bm{X}$ such that
\begin{equation}
f(\bm{X}) = \|F(\bm{X})\|_{\bm{W}}^2 = F(\bm{X})^\top \bm{W} F(\bm{X})
\label{NLLSSolution}
\end{equation}
is minimised. This formulation can be solved by the Gauss-Newton method. The update vector $\bm{\Delta}$ in each iteration is the solution to
\begin{equation}
\bm{J}^\top \bm{W} \bm{J} \bm{\Delta} = -\bm{J}^\top \bm{W} F(\bm{X})
\end{equation}
where \( \bm{J} \) is the Jacobian \({\partial F(\bm{X})}/{\partial X}\).

\subsubsection{Analytical Jacobian}

We now derive the analytical Jacobian of our NLLS formulation to accelerate algorithm convergence and enhance robustness.

The Jacobian matrix $\bm{J}$ consists of three parts: the Jacobian of the depth constraint term w.r.t. the poses $\bm{J}_{{P}}$, the Jacobian of the depth constraint term w.r.t. the depth map $\bm{J}_{{D}} $, and the Jacobian of the smoothing term w.r.t. the depth map $\bm{J}_S$. 

For the depth constraint term $f^D(\bm{X})$, the Jacobian w.r.t. the  pose $\bm{X}^r_i$ is give by 
\begin{equation}
\begin{aligned}
\bm{J}_{{P}} &= \frac{\partial \left( [\bm{p'}_{ij}]_z - \bm{D}([\bm{p'}_{ij}]_{xy}/s) \right)}{\partial \bm{X}_i^r} \\
& =  \frac{\partial [\bm{p'}_{ij}]_{z}}{\partial \bm{X}_i^r} -\frac{\partial \bm{D}([\bm{p'}_{ij}]_{xy}/s)}{\partial([\bm{p'}_{ij}]_{xy}/s)} \cdot \frac{\partial [(\bm{p'}_{ij}]_{xy}/s)}{\partial \bm{X}_i^r}.
\end{aligned} \label{eq_JP}
\end{equation}
We can first calculate $\partial (\bm{p'}_{ij}/s)/{\partial \bm{X}_i^r}$ as
\begin{equation}
\frac{\partial (\bm{p'}_{ij}/s)}{\partial \bm{X}_i^r} 
= \left[ \frac{\partial (\bm{p'}_{ij}/s)}{\partial \bm{t}_i^r}, \frac{\partial (\bm{p'}_{ij}/s)}{\partial \bm{\theta}_i}\right] = \frac{1}{s} \cdot \left[\bm{I}_{3*3},\bm{R}_i^\prime {\bm{p}_{ij}}\right],
\end{equation}
where $\bm{R}_i^\prime$ represents the derivative of $\bm{R}_i$ w.r.t. the orientation. Then we can obtain
\begin{equation}
\begin{aligned}
&\frac{\partial ([\bm{p'}_{ij}]_{xy}/s)}{{\partial \bm{X}_i^r}} = \frac{1}{s} \cdot [\bm{I}_{3*3},\bm{R}_i^\prime {\bm{p}_{ij}}]_{xy},\\
    &\frac{\partial [\bm{p'}_{ij}]_{z}}{{\partial \bm{X}_i^r}} = [\bm{I}_{3*3},\bm{R}_i^\prime {\bm{p}_{ij}}]_{z},
\end{aligned}
\end{equation}
where $[\cdot]_{xy}$ and $[\cdot]_z$ are defined in (\ref{eq_xyz}). 

In (\ref{eq_JP}), $\partial \bm{D}([\bm{p'}_{ij}]_{xy}/s)/\partial({[\bm{p'}_{ij}]_{xy}}/s)$ can be considered as the gradient of the depth map at point $[\bm{p'}_{ij}]_{xy}/s$, which can be approximated by the bilinear interpolation of the gradients of the depth at the four adjacent grid ${\nabla} \bm{D}(\bm{d}_{(m,n)}),\cdots,{\nabla} \bm{D}(\bm{d}_{(m+1,n+1)})$ around $[\bm{p'}_{ij}]_{xy}$ as 
\begin{equation} 
\begin{aligned}
    \dfrac{\partial \bm{D}([\bm{p'}_{ij}]_{xy}/s)}{\partial ([\bm{p'}_{ij}]_{xy}/s)} = \bm{H}\left[\begin{array}{l}
\nabla\bm{D}(\bm{d}_{m,n}) \\
\nabla\bm{D}(\bm{d}_{m+1,n}) \\
\nabla\bm{D}(\bm{d}_{m,n+1}) \\
\nabla\bm{D}(\bm{d}_{m+1,n+1})
\end{array}
    \right]
\end{aligned}
\end{equation} 
where the gradient of depth map $\bm{D}$ at all the grid ${\nabla} \bm{D}$ can be easily calculated from the depth map.

The Jacobian of depth constraint term w.r.t. all the grids of depth map $\bm{J}_D$ can be easily calculated as

\begin{equation}
\begin{aligned}
\bm{J}_D & =  -\frac{\partial \bm{D}([\bm{p'}_{ij}]_{xy}/s)}{\partial \left[\cdots, \bm{D}(\bm{d}_{m,n}),\cdots, \bm{D}(\bm{d}_{m+1,n+1}),\cdots) \right]^\top}
\\
&= -[0,\cdots,\bm{H},\cdots,0].
\end{aligned}
\end{equation} 

Finally, it is easy to find that the Jacobian of the smoothing term w.r.t. the depth map $\bm{J}_S$ is equal to a constant matrix that consists of $1$, $-1$, and $0$.

\begin{table*}[thp]
\vspace{1.8mm}
\caption{Poses Accuracy of Different Methods Evaluated by Laboratory Dataset}
\label{tab:Indoor_results}
\setlength{\tabcolsep}{17pt} 
\renewcommand{\arraystretch}{1.2} 
\begin{tabular}{clcccccc}
\toprule
\textbf{Scene} & \textbf{Metric} & \textbf{TEASER} & \textbf{T+ICP} & \textbf{T+BA} & \textbf{T+PGO} & \textbf{BALM2} & \textbf{Ours} \\ \midrule \\[-3ex]

\multirow{4}{*}{1} 
& MAE((Trans/m)) & 0.0141 & 0.0113 & \textcolor{blue}{0.0083} & 0.0135 & 0.0333 & \textcolor{red}{0.0068} \\[-0.5ex]
& RMSE((Trans/m)) & 0.0202 & 0.0152 & \textcolor{blue}{0.0139} & 0.0206 & 0.0527 & \textcolor{red}{0.0099} \\[-0.5ex]
& MAE(Rot/rad) & 0.0239 & \textcolor{red}{0.0118}  & 0.0159 & 0.0215 & 0.0446 & \textcolor{blue}{0.0146} \\[-0.5ex]
& RMSE(Rot/rad) & 0.0349 & \textcolor{red}{0.0159} & 0.0238 & 0.0345 & 0.0944 & \textcolor{blue}{0.0217} \\ \midrule \\[-3ex]

\multirow{4}{*}{2} 
& MAE((Trans/m)) & \textcolor{blue}{0.0114} & 0.0124 & 0.0169 & 0.0115 & 0.0413 & \textcolor{red}{0.0075} \\[-0.5ex]
& RMSE((Trans/m)) & 0.0144 & 0.0165 & 0.0292 & \textcolor{blue}{0.0133} & 0.0655 & \textcolor{red}{0.0092} \\[-0.5ex]
& MAE(Rot/rad) & \textcolor{red}{0.0111} & 0.0119 & 0.0295 & 0.0229 & 0.1167 & \textcolor{blue}{0.0116} \\[-0.5ex]
& RMSE(Rot/rad) & 0.0145 & \textcolor{blue}{0.0153} & 0.0479 & 0.0265 & 0.1759 & \textcolor{red}{0.0151} \\ \midrule \\[-3ex]

\multirow{4}{*}{3} 
& MAE((Trans/m)) & 0.0392 & 0.0482 & 0.0579 & 0.0643 & \textcolor{blue}{0.0377} & \textcolor{red}{0.0139} \\[-0.5ex]
& RMSE((Trans/m)) & 0.6266 & 0.0655 & 0.1260 & 0.0915 & \textcolor{blue}{0.0638} & \textcolor{red}{0.0194} \\[-0.5ex]
& MAE(Rot/rad) & \textcolor{blue}{0.0302} & 0.2421 & 0.2093 & 0.2137 & 0.0327 & \textcolor{red}{0.0141} \\[-0.5ex]
& RMSE(Rot/rad) & \textcolor{blue}{0.0384} & 0.6266 & 0.5797 & 0.3451 & 0.0524 & \textcolor{red}{0.0211} \\\midrule

\end{tabular}
\end{table*}

\section{EXPERIMENTS}

\subsection{Baseline}
To evaluate our method's effectiveness, we compare it against five state-of-the-art approaches:

1) Pairwise registration methods: TEASER \cite{yang2020teaser} is one of the most robust and efficient pairwise registration methods. To further enhance the accuracy of TEASER, we incorporate ICP into TEASER for fine registration, denoted as T+ICP. To extend pairwise registration to the multiview registration problem, we apply sequential registration across multiple frames to estimate their poses and refine the final global point cloud.

2) Integrating TEASER with batch optimisation: Sequential multiple executions of TEASER in the multiview point cloud registration task inevitably lead to pose error accumulation, resulting in an inconsistent global point cloud map. To address this issue, additional batch optimisation is required to enhance global accuracy across multiple frames. We integrate TEASER with two batch optimisation strategies:
\begin{itemize}
    \item T+PGO: We apply pose graph optimisation to refine the global alignment, treating the results from TEASER as relative pose measurements.
    \item T+BA: We extract feature points from point clouds to formulate a feature-based bundle adjustment problem, using the results from sequential TEASER execution as the initial guess.
\end{itemize}

3) Bundle adjustment methods based on alternative feature representations: BALM2 \cite{liu2023efficient} is the state-of-the-art method in this category, leveraging planar features rather than feature points, as used in T+BA, to construct bundle adjustment.

\begin{figure}[t!]
    \centering
    \vspace{2mm}
    \begin{subfigure}{0.48\linewidth}
        \centering
        \includegraphics[width=4cm, height=2.5cm]{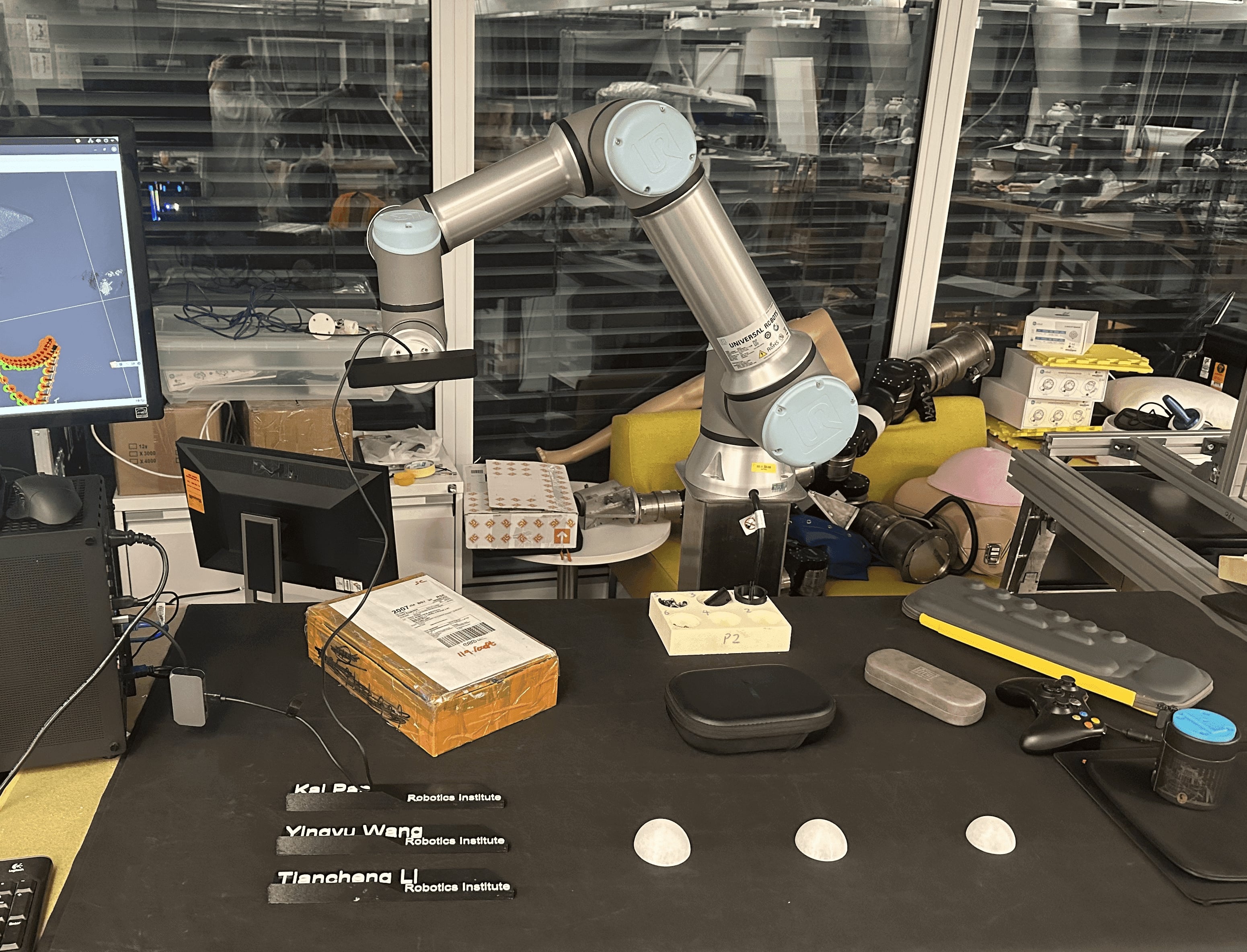}
        \caption{Laboratory Environment}
        \label{fig:left}
    \end{subfigure}
    \hfill
    \begin{subfigure}{0.48\linewidth}
        \centering
        \includegraphics[width=4cm, height=2.5cm]{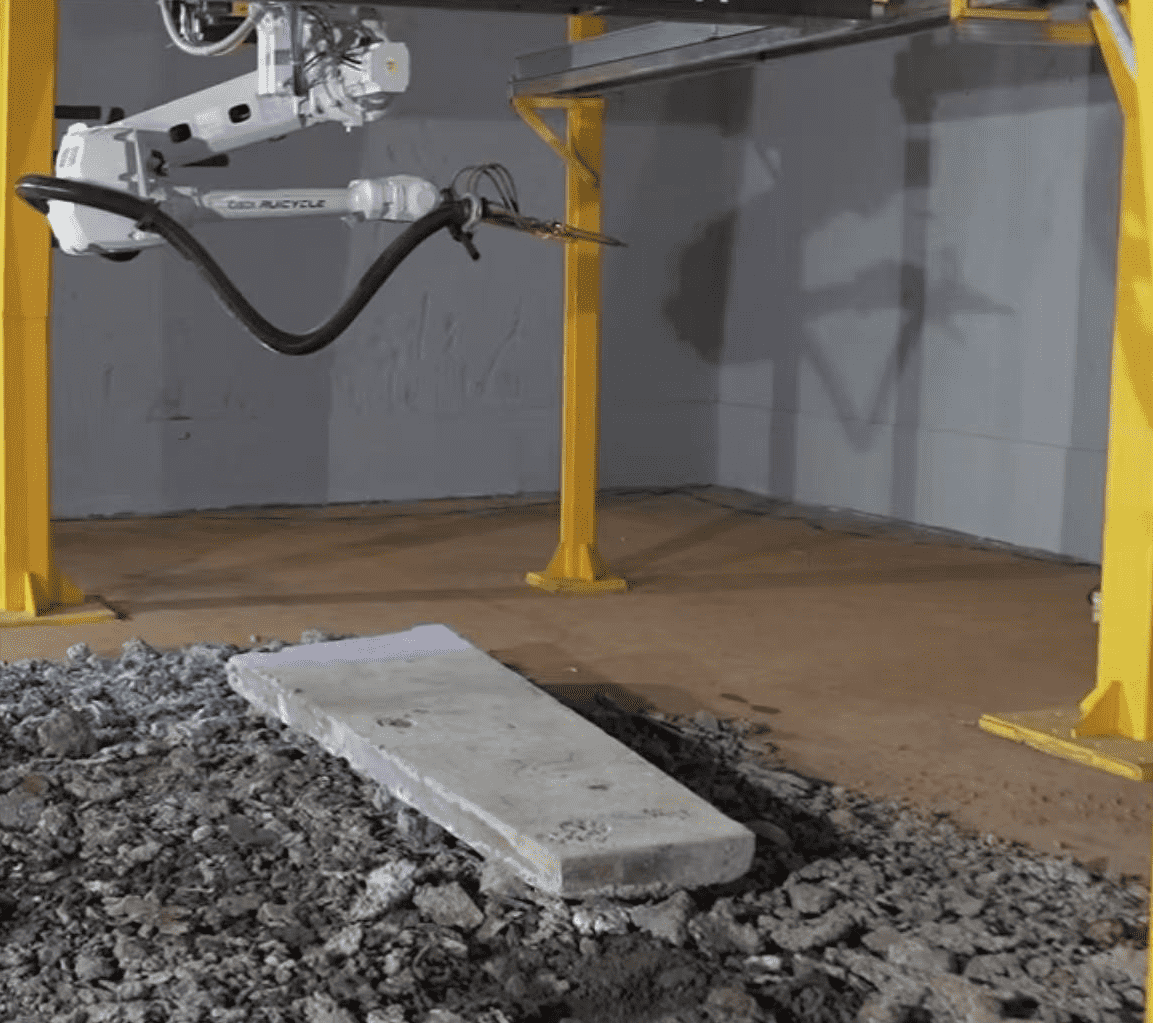}
        \caption{Industrial Environment}
        \label{fig:right}
    \end{subfigure}
    \caption{The environmental setup for data collection includes both laboratory and industrial scenes. The laboratory environment measures 2.4$*$1.2$*$0.7 (m), and industrial environment measures 7.0$*$6.0$*$3.0 (m).}
    \label{fig:environments}
\vspace{-1.5em}
\end{figure}

\begin{figure*}[th!]
\centering
\includegraphics[width=\textwidth]{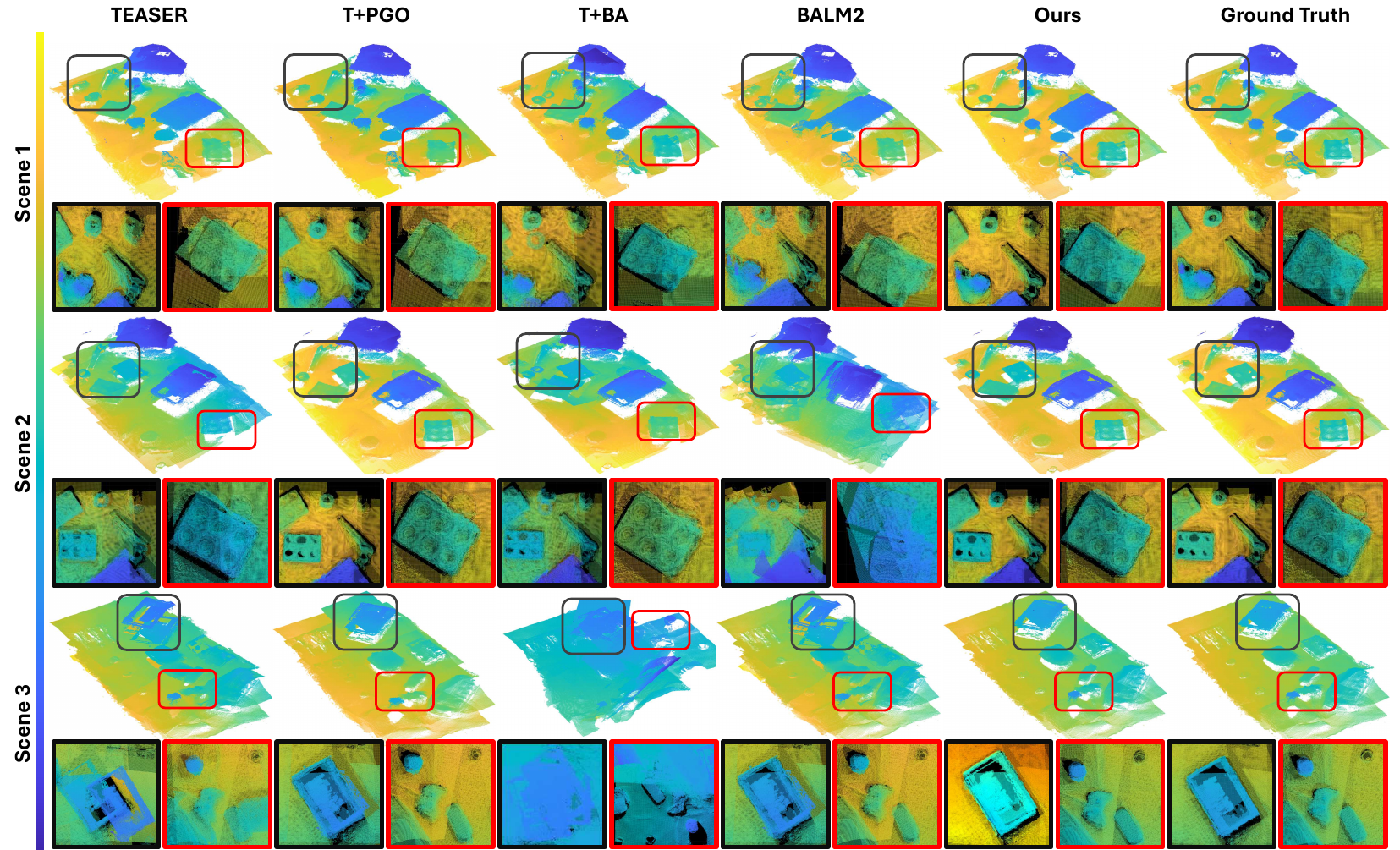}
\caption{Global point clouds registered by different methods using the laboratory dataset. The areas highlighted by red and black boxes show detailed figures to enhance the comparison between our and other methods.}
\label{fig:Indoor}
\vspace{-1.5em}
\end{figure*}

\subsection{Experimental Datasets and Setup}
Our experiments involve two distinct self-collected datasets: a laboratory dataset, captured in a controlled laboratory environment with three scenes (Scene 1-3) and a industrial dataset, recorded in a less structured environment, including three scenes (Scene 4-6) as well. Fig. \ref{fig:environments} illustrates the experimental setups for both environments.

\subsubsection{Laboratory dataset}
We collected the dataset with reliable ground truth using a ZED 2 camera mounted on a UR16 robotic arm. The depth accuracy of the camera is less than 1\% of the measured depth, and the pose translation errors of UR16 are within 0.05 mm. 
\begin{figure*}[th!]
\centering
\includegraphics[width=\textwidth]{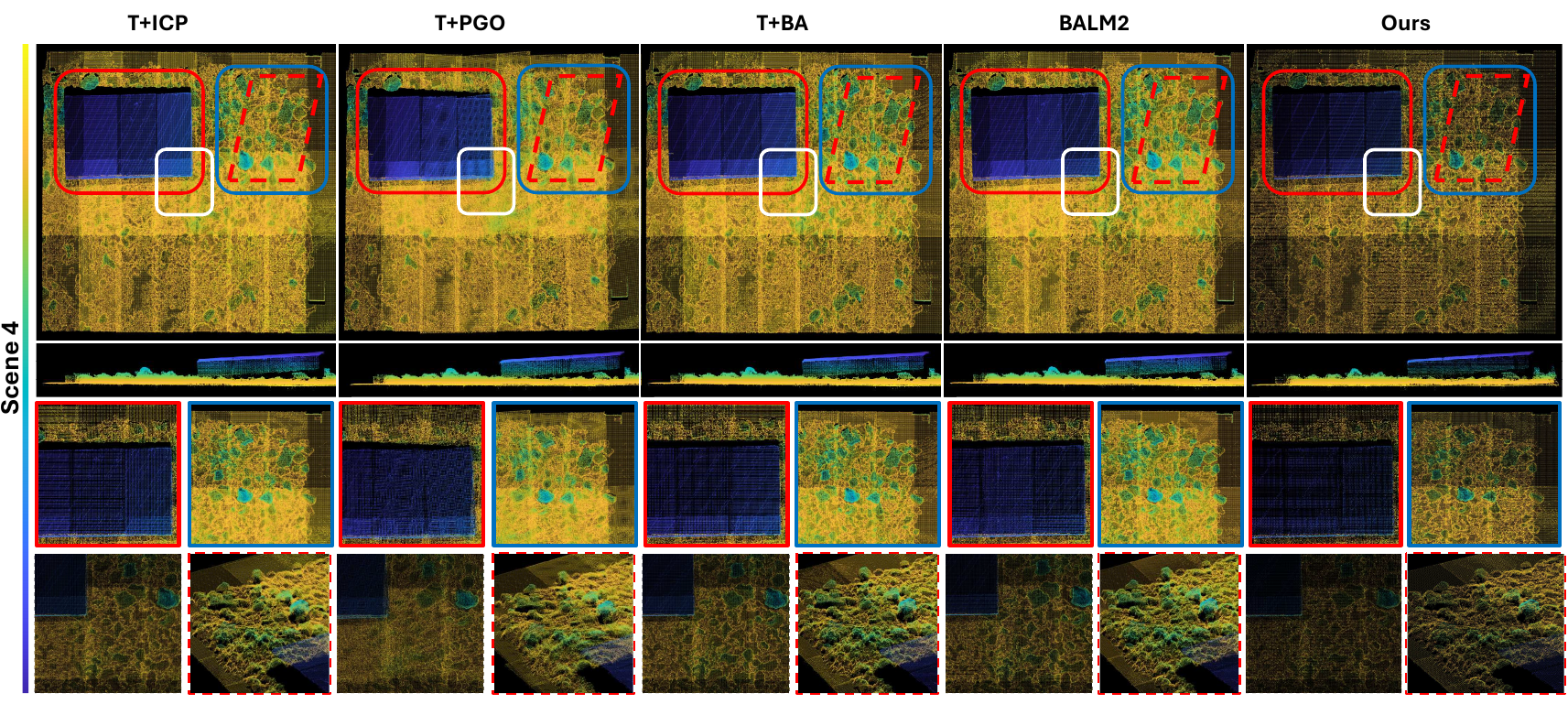}
\caption{Comparison of performance among different methods applied to industrial dataset Scene 4.}
\label{fig:Cu}
\vspace{-1em}
\end{figure*}
To obtain accurate ground truth poses, we captured multiple images of a checkerboard from different positions and orientations while simultaneously recording the end-effector poses of the robotic arm at each position. The hand-eye calibration algorithm \cite{park1994robot} was then applied to compute the transformation between the camera and the end-effector frame, ensuring accurate alignment with the robotic arm's frame.  In addition, the ground truth of the global point cloud is obtained by transforming local point clouds into a global coordinate system using known ground truth poses of depth camera. 

\subsubsection{Industrial dataset}
The industrial experiments involved more complex experiments. The setup featured a truss system securely mounted with a structured-light camera, SEIZET SP1000, which provides a depth accuracy of 0.32 mm at a working distance of  3000 mm. The system moves in the X-Y direction only, which is driven by two motors, ensuring a constant height throughout data collection. 

The objects captured in this dataset included crushed stone, steel slag, and various types of scrap. The dataset consists of industrial scenes characterised by more complex terrain and less controlled conditions to evaluate performance across different geometric structures. Including a steel slab with a planar surface, a steel coil with a curved surface, and a medium-thick plate with multiple planar surfaces, while keeping the ground features unchanged.

Unlike the laboratory dataset, this dataset does not include ground truth. To facilitate the evaluation of algorithm accuracy, three markers on the floor were strategically placed and fixed within Scenes 4-6. Specifically, one marker was placed at the origin of the global frame, while the other two were positioned at the farthest locations along the X-Y direction. The known distances between these markers serve as a reference for accuracy evaluation.

\begin{figure*}[th!]
\centering
\includegraphics[width=\textwidth]{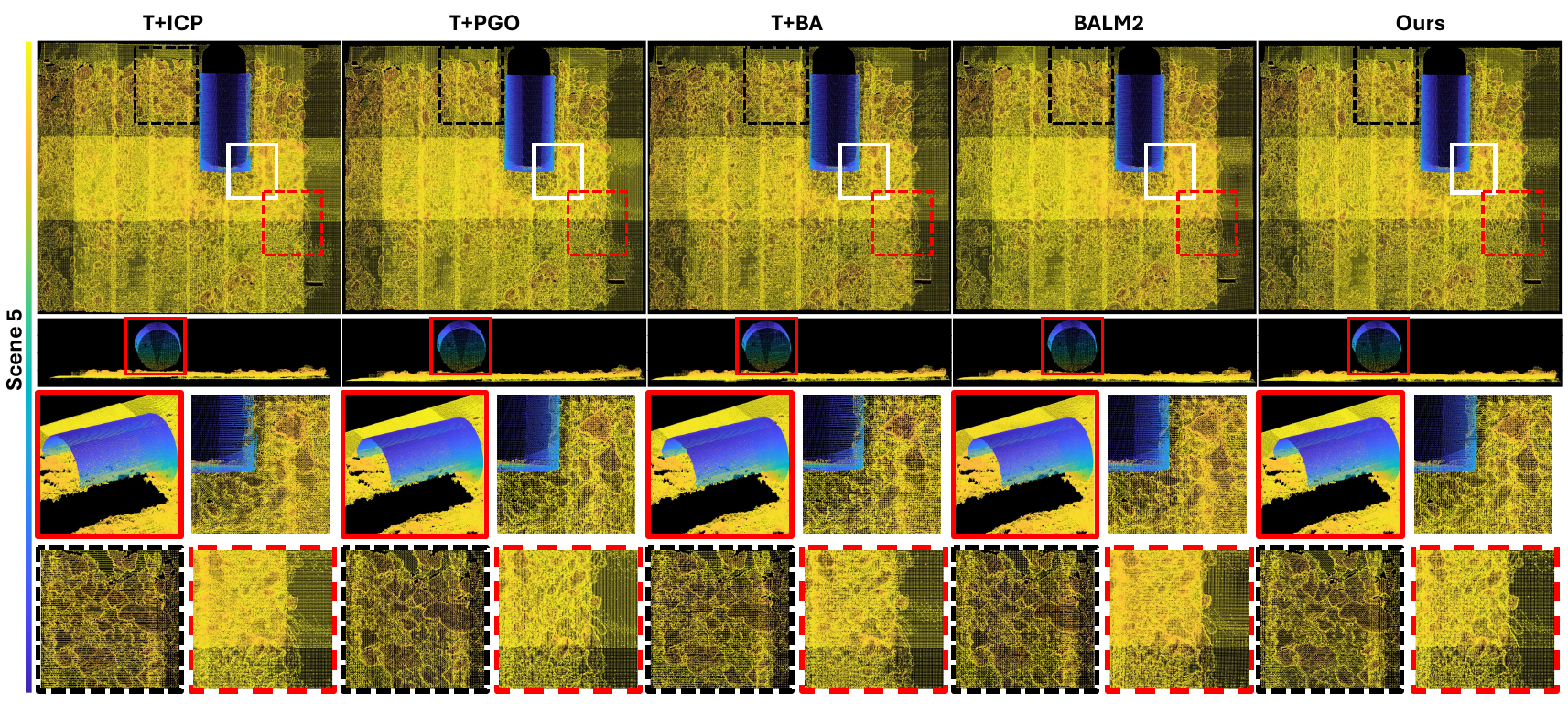}
\caption{
Comparison of performance among different methods applied to industrial dataset Scene 5. }
\label{fig:Cy}
\vspace{-1.3em}
\end{figure*}

\subsection{Evaluation of Pose Accuracy}\label{Accuracy}

We first quantitatively evaluate the accuracy of the estimated poses using the laboratory dataset, where ground truth of poses are available. Table \ref{tab:Indoor_results} presents the quantitative accuracy results across different state-of-the-art methods. We evaluate pose accuracy using Mean Absolute Error (MAE) and Root Mean Squared Error (RMSE) for both translation (in metres) and orientation (in radians). The best-performing values are highlighted in red, and the second-best in blue. 
\begin{figure*}[th!]
\centering
\includegraphics[width=\textwidth]{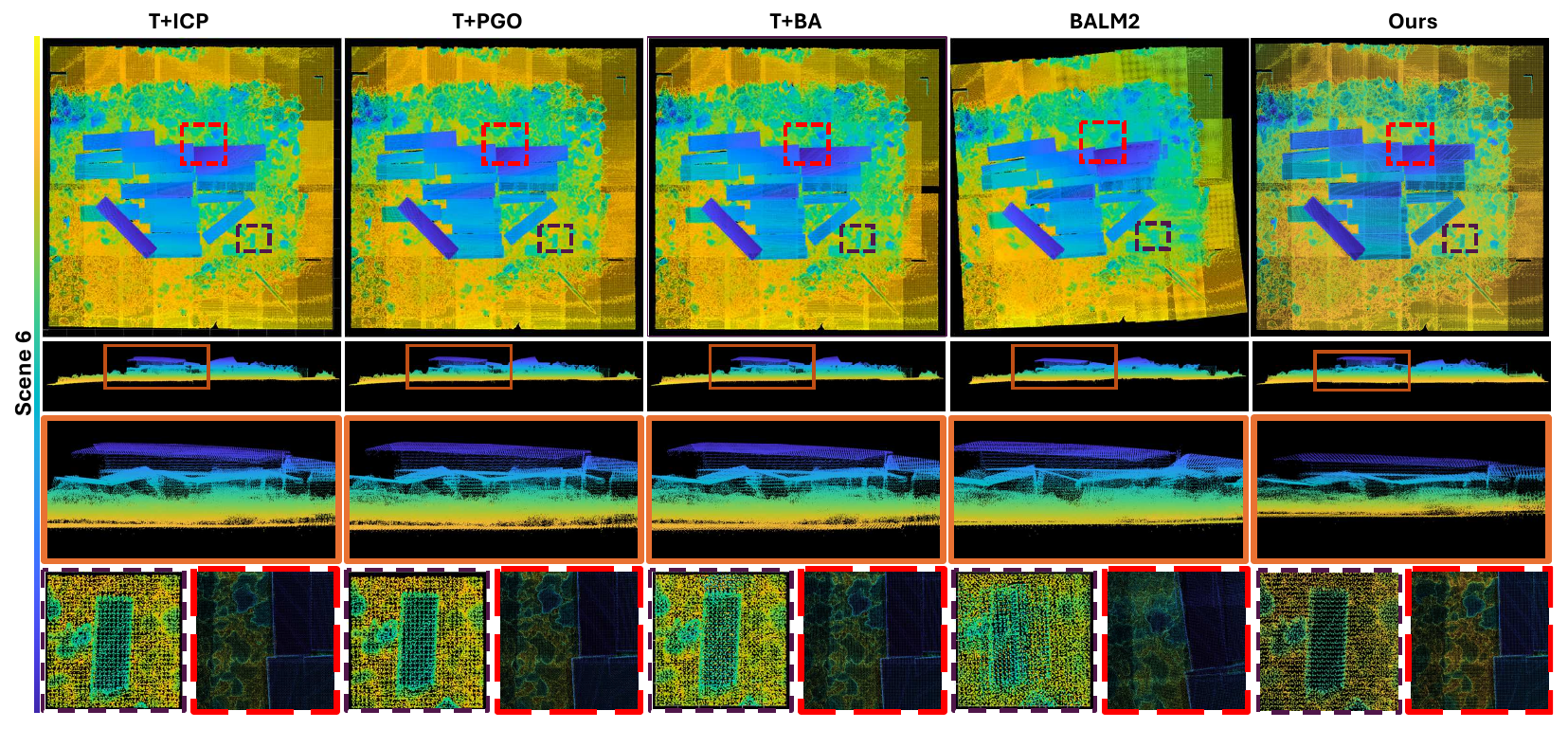}
\caption{
Comparison of performance among different methods applied to industrial dataset Scene 6.}
\label{fig:P}
\captionsetup[figure]{skip=8pt}
\vspace{-1.3em}
\end{figure*}

As shown in Table \ref{tab:Indoor_results}, our method achieves the lowest errors across the majority of metrics. In Scene 1, while T+ICP attains the best orientation accuracy, it performs significantly worse in translation against ours. A similar trend is observed in Scene 2, where TEASER achieves better orientation accuracy but exhibits higher translation errors than ours. This demonstrates that some methods excel in one metric but fail to maintain overall accuracy. 

A noteworthy observation is that T+BA does not perform well on the laboratory dataset and even underperforms compared to TEASER on several metrics. This is partly because the dataset covers a small scene size with a limited number of frames, resulting in a relatively low accumulation of errors from the sequential execution of TEASER. Consequently, the impact of global optimisation in T+BA is less significant. More importantly, the laboratory dataset contains many regions with repetitive textures, making it difficult for T+BA to establish accurate data association between frames. This challenge directly affects the effectiveness of feature-based bundle adjustment, ultimately reducing its registration accuracy. These results highlight the limitations of T+BA in environments with repetitive structures, where unreliable feature correspondences negatively impact its performance. In contrast, our method does not require feature extraction or explicit data association, allowing it to avoid these limitations. As a result, it remains robust in environments with repetitive textures, maintaining high registration accuracy regardless of scene structure.
Additionally, BALM2 does not achieve good results on the laboratory dataset for most metrics. This is because BALM2 relies on planar feature extraction and constructs planar constraints to formulate the bundle adjustment problem. However, in this scenario, many objects are not entirely planar, limiting its effectiveness. In contrast, our approach eliminates the need for feature extraction by associating the point cloud with the global depth map, effectively avoiding this issue.

\subsection{Evaluation of Global Point Cloud Quality}

We first qualitatively assess the map quality by visualising the registration results of the 3D global point cloud using both the laboratory dataset and the industrial dataset.

The 3D global point clouds registered by different methods using the laboratory dataset are visualised in Fig. \ref{fig:Indoor}. The areas highlighted by red and black rectangles illustrate that our method achieves superior results compared to the others. It can be observed that our method produces a reconstruction closest to the ground truth, with clearer and sharper contour details and well-defined holes. In contrast, the global point cloud registrations produced by TEASER and T+PGO exhibit fuzzy edges and alignment errors across regions. This demonstrates two key issues: first, the sequential execution of TEASER in multiview registration leads to error accumulation, impacting overall accuracy. Second, PGO focuses solely on pose refinement while ignoring the global map, resulting in registered point cloud maps with poorer detail preservation. These findings highlight the importance of jointly optimising both poses and the global map in multiview registration to achieve a more accurate and consistent reconstruction. While BALM2 generally performs well, it struggles in Scene 2 due to the scarcity of high-quality planar features. Notably, T+BA outperforms TEASER in Scene 1 and Scene 2 but fails in Scene 3, likely due to incorrect data association. This highlights the limitations of traditional feature-based BA approaches in complex scenes where feature correspondences become unreliable.

\begin{table}[htbp]
\caption{Marker Distances Errors on Industrial dataset}
\label{industrialDistance}
\setlength{\tabcolsep}{4.8pt} 
\renewcommand{\arraystretch}{1.2}
\begin{tabular}{@{}ccccccc@{}}
 \midrule
\textbf{Scene} & \textbf{Direction} & \textbf{TEASER} & \textbf{T+PGO} & \textbf{T+BA} & \textbf{BALM2} & \textbf{Ours}  \\ 
\midrule \\[-2ex] 
\multirow{2}{*}{4}  
& Y (m) & 0.0479 & \textcolor{red}{0.0198} & 0.0327 &                 0.0339 & \textcolor{blue}{0.0300} \\
& X (m) & 0.0343 & \textcolor{red}{0.0059}  & 0.0607 & 0.0534 & \textcolor{blue}{0.0152}   \\[1ex]\midrule 

\multirow{2}{*}{5} 
& Y (m) & 0.0369 & \textcolor{blue}{0.0246} & 0.0414 & 0.0317 & \textcolor{red}{0.0226}  \\ 
& X (m) & 0.0429 & 0.0413 & 0.0530 & \textcolor{blue}{0.0371} & \textcolor{red}{0.0289} \\[1ex] \midrule 

\multirow{2}{*}{6}
& Y (m) & 0.0768 & \textcolor{blue}{0.0105} & 0.0139 &                 0.0451 & \textcolor{red}{0.0091}   \\  
& X (m) & \textcolor{red}{0.0186} & 0.0096 & 0.0822 & {0.0536} & \textcolor{blue}{0.0367} \\[1ex] \midrule 

\end{tabular}
\vspace{-1em}
\end{table}

We also qualitatively evaluate map quality using the industrial datasets characterised by less structured environments. As shown in Fig. \ref{fig:Cu}, Fig. \ref{fig:Cy} and Fig. \ref{fig:P}, our method yields a darker global map, indicating higher registration accuracy in overlaps and a more compact reconstruction. Detailed views show sharp contours, preserving fine structures and complex details like scrap geometries and irregular slag edges. For example, in Fig. \ref{fig:Cu}, the red dashed box highlights a steel slag region with distinct colour gradients and sharp edges, while the rectangular billet’s upper edge is precisely aligned. In contrast, T+ICP and T+PGO suffer from noticeable blurring and splicing mismatches, with T+PGO showing significant edge misalignment. Similarly, in Fig. \ref{fig:Cy}, the white solid box highlights the edge of the steel coil, where T+ICP shows similar misalignment. Although T+BA and BALM2 improve detail preservation, localised distortions and blurring remain, as highlighted by the blue boxes and red dashed in Fig. \ref{fig:Cu} and Fig. \ref{fig:Cy}. In contrast, the global point cloud at the top and the deep purple dashed box in Fig. \ref{fig:P} show that T+BA has one misaligned frame, while BALM2 suffers even more misalignment.

Finally, we leverage prior information on marker distances in the industrial dataset to compute surface distance errors, providing a quantitative measure of global point cloud registration quality. As shown in Table \ref{industrialDistance}, our method achieves either the best or second-best performance across all metrics. While T+PGO and TEASER slightly outperform our method in certain metrics, the qualitative results in Fig. \ref{fig:Cu}, Fig. \ref{fig:Cy} and Fig. \ref{fig:P} reveal inconsistencies in the overlaps. This suggests that these frames may suffer from large orientation errors or frames without markers may have inaccurate poses. Therefore, our method demonstrates the most reliable global point cloud registration performance on the industrial dataset.

\subsection{Time Consumption}
We further compare the registration runtime with BALM2, T+BA, and T+PGO using Scene 6, which consists of 21 frames of point clouds. For our method, under a resolution of 0.05 m, the number of optimisation iterations until convergence is 15, and the runtime per iteration is approximately 3–5 seconds. In comparison, BALM2 completes the optimisation in around 5 seconds, while T+BA requires 45 seconds in total. Although PGO's pose-only optimization is inherently fast, obtaining multi-frame relative poses using TEASER is significantly more time-consuming, averaging 10 seconds per relative pose computation.

\section{Conclusion}
This paper formulates the multiview point cloud registration task as a correspondence-free bundle adjustment problem. The key novelty lies in implicitly associating point clouds with a depth map, eliminating the need for explicit feature extraction and data association. As a result, our method avoids the errors commonly encountered in feature-based BA approaches due to insufficient feature extraction or incorrect data association in low-texture, repetitive-texture, or highly unstructured scenarios. We evaluate our method on two self-collected datasets, covering both laboratory and industrial environments, and demonstrate that it outperforms state-of-the-art algorithms, particularly in challenging scenarios where feature extraction and association across multiple frames are difficult.

\bibliographystyle{IEEEtran}
\bibliography{IEEEtranBST/Bibliography}

\end{document}